\documentclass[runningheads]{llncs}
\usepackage{geometry}

\usepackage{graphicx}
\usepackage{amsmath}
\usepackage{amssymb}
\usepackage{xspace}
\usepackage{xcolor}
\usepackage{dsfont}
\usepackage{rotating}
\usepackage{makecell}
\usepackage{multirow}
\usepackage{algpseudocode,algorithm,algorithmicx}
\usepackage{subcaption}

\newcommand\eps{\varepsilon}

\newcommand{\onemax}{\textsc{OneMax}\xspace}

\newcommand{\hottopic}{\textsc{HotTopic}\xspace}
\newcommand{\BV}{\textsc{BV}\xspace}
\newcommand{\DBV}{\textsc{DBV}\xspace}

\newcommand{\oea}{${(1+1)}$-EA\xspace}

\newcommand{\olea}{$(1 + \lambda)$-EA\xspace}

\newcommand{\moea}{$(\mu + 1)$-EA\xspace}
\newcommand{\moga}{$(\mu + 1)$-GA\xspace}
\newcommand{\ollga}{$(1 + (\lambda,\lambda))$-GA\xspace}

\newcommand{\folea}{$(1 + \lambda)$-fEA\xspace}
\newcommand{\fmoea}{$(\mu + 1)$-fEA\xspace}

\usepackage{orcidlink}
\begin{document}
\title{Empirical Analysis of the Dynamic Binary Value Problem with IOHprofiler}
\titlerunning{Analysis of Dynamic Binary Value Problem}
%
\author{Diederick Vermetten\inst{1}\orcidlink{0000-0003-3040-7162}, 
Johannes Lengler\inst{2}\orcidlink{0000-0003-0004-7629},
Dimitri Rusin\inst{3}\orcidlink{0009-0007-3316-0480},
Thomas B\"ack\inst{1}\orcidlink{0000-0001-6768-1478},
Carola Doerr\inst{3}\orcidlink{0000-0002-4981-3227}
}
\institute{
LIACS, Leiden University, Leiden, The Netherlands
\and 
ETH Z\"urich, Z\"urich, Switzerland
\and 
Sorbonne Universit\'e, CNRS, LIP6, Paris, France
}
\authorrunning{Vermetten, Lengler, Rusin, B\"ack, Doerr}
%

%
\maketitle              
\begin{abstract}
Optimization problems in dynamic environments have recently been the source of several theoretical studies. One of these problems is the monotonic Dynamic Binary Value problem, which theoretically has high discriminatory power between different Genetic Algorithms. 
Given this theoretical foundation, we integrate several versions of this problem into the IOHprofiler benchmarking framework. Using this integration, we perform several large-scale benchmarking experiments to both recreate theoretical results on moderate dimensional problems and investigate aspects of GA's performance which have not yet been studied theoretically. 
Our results highlight some of the many synergies between theory and benchmarking and offer a platform through which further research into dynamic optimization problems can be performed.   

\keywords{Evolutionary Algorithms  \and Benchmarks \and Dynamic Environment \and Dynamic Binary Value.}
\end{abstract}

\section{Introduction}
Evolutionary and genetic algorithms (EAs, GAs) are an important family of randomized optimization heuristics. In order to better understand the behavior of these algorithms, we should take advantage of the wide range of perspectives from which they have been studied, including theoretical runtime analysis, systematic benchmarking studies, and practical experience. While each of these domains offers its own viewpoint, these should not stay isolated but rather benefit from and strengthen each other. 

In the last years, there has been a substantial amount of theoretical work on two seemingly unrelated situations: optimization of monotonic functions~\cite{lengler2019general}, and optimization in dynamic linear environments~\cite{lengler2018one,Janett2023two} like dynamic binary value~\cite{lengler2022large}. For both of these topics, theoretical analysis shows that even among simple EAs, performance can dramatically differ. Hence, these situations have high discriminative power between different EAs. To give a flavour of the theoretical results, we highlight two specific results here:
\begin{enumerate}
    \item The \oea with mutation rate $1/N$ finds the optimum of the dynamic binary value environment in time $O(N\log N)$, while the same algorithm with mutation rate $2/N$ needs exponential time~\cite{lengler2018one}.
    \item The \oea with mutation rate $1/N$ finds the optimum of every monotonic function in time $O(N\log^2 N)$~\cite{lengler2019does}, while the \moea with the same mutation rate and large population size $\mu$ needs exponential time on some hard instances~\cite{lengler2019exponential}.
\end{enumerate}
These results illustrate the strengths, but also the limitations of theoretical analyses. They can identify interesting differences between similar algorithms, and may be able to prove results on whole classes of benchmarks like all monotonic functions. Moreover, they are able to provide a deep understanding of the given situation. On the other hand, they are often limited to rather simple algorithms, and to asymptotic statements like ``$O(N \log N)$'' or ``exponential''. Thus they leave important gaps that can be filled with systematic benchmarking studies.

In this paper we provide such a benchmarking study for two aforementioned situations of monotonic and of dynamic linear environments. To this end, we have several contributions:
\begin{itemize}
    \item We develop practical variants of the theoretical benchmarks that have been used for runtime analyses. This requires non-trivial modifications, as the theoretical variants are not suited for efficient evaluations, see Section~\ref{sec:theory-benchmarks}.
    \item We have integrated them into IOHexperimenter~\cite{iohexp},
    a module of the open-source IOHprofiler~\cite{doerr2019benchmarking} project that allows general access to the new benchmarks in a framework where they can be easily used to test other iterative optimization algorithms. Part of this integration is a generic extension of IOHexperimenter to support dynamic environments.
    \item We run a systematic evaluation of a large class of algorithms on the new benchmarks, identifying the parameters which are generally most crucial for the performance on these benchmarks.
    \item We investigate several of the theoretical results in a wider context, asking how big the effects are for small dimensionality $N$, how sensitive they are for other parameter settings and how well they transfer to modifications of the algorithms. 
\end{itemize}

Since this paper brings together many different benchmarking aspects, we discuss them in more detail in the following sections.

\subsection{Dynamic environments}

As mentioned, we have extended the IOHprofiler framework~\cite{doerr2019benchmarking} to support dynamic environments, i.e., environments in which the fitness function $f$ can change over time. In our experiments, whenever the environment changes, we re-evaluate the population with respect to the new environment, to avoid comparing fitnesses from different environments with each other. However, the implementation also supports keeping the old fitness values.

Dynamic environments can occur in various ways through a range of applications~\cite{branke2012evolutionary}. 
 For example, part of the problem description may be uncertain or become only available over time~\cite{nguyen2012evolutionary,neumann2020analysis}. This work is based on dynamic linear functions~\cite{lengler2018one,lengler2022large,Janett2023two} and the dynamic binary value function~\cite{lengler2021runtime,lengler2022large}, which were motivated by shifting training data for a machine learning system, especially in the context of co-evolution~\cite{lengler2022large}. 

\subsection{Theory-inspired benchmarks}\label{sec:theory-benchmarks}

In recent years, there has been increasing interest and demand for composing a problem suite with theoretical benchmarks that are suitable for empirical testing. We provide a class of such benchmarks, based on two ideas from theory: dynamic binary value for dynamic environments and the \hottopic function as an instance of a hard monotonic function. 

The dynamic binary value function \DBV is based on the static linear binary value function $\BV:\{0,1\}^N \to \mathbb{R}; \BV(x) = \sum_{i=1}^N 2^{i-1}x_i$, which computes for a bit string $x$ the integer encoded by $x$ in binary representation. In the dynamic setting, we use the same set of weights, but shuffle them via a random permutation. Note that the same permutation is applied for all strings that are evaluated in the same environment. Hence, every environment is given by a linear function with positive weights. In particular, all environments share the all-one string as their common global optimum, which gives the optimization process a clear objective: it should produce solutions which are as close as possible to the invariant global optimum. Moreover, the functions are never deceptive in any coordinate: whenever we flip a zero-bit into a one-bit, then the fitness increases in all possible environments. This ensures that the benchmark is feasible and can be solved efficiently by some algorithms and parameter configurations, but not by all. This yields the high discriminatory power of the benchmark. 

Practically, the $\BV$ function can not easily be implemented because the weights are so large that they cause numerical problems. We do provide an implementation via a lexicographic comparison, but since this requires an approach that is not compatible with the usual framework of fitness functions, it is less well-suited as a module to be used by other researchers. Instead, we use the observation from~\cite{lengler2022large} that \DBV can be obtained in the limit from dynamic linear functions where we draw all weights independently from some heavy-tailed distribution. 
In this work, we consider the Uniform, Power-of-two (with a maximum power) and capped Pareto distributions, and compare them to the version based on lexicographic comparison. 

Moreover, our implementation leaves the freedom to either change the environment every generation, or to change it less frequently. This also gives a way to test a function similar to the \hottopic function introduced in~\cite{lengler2018drift} (named \hottopic in~\cite{lengler2019general}). This function has served an important role as theoretical benchmark~\cite{lengler2019general,lengler2019exponential}, but it is forbiddingly slow to evaluate practically. However, it has been pointed out in~\cite{lengler2021runtime} that \hottopic may be approximated by a dynamic linear function in which the environment changes every $\eps n$ generations, for some constant $\eps$. 

Thus, we provide several practical implementations of theoretical benchmarks. We empirically validate our implementations in Section~\ref{sec:results} by testing whether previous theoretical analyses can be recovered empirically, and find generally a good qualitative agreement.

\subsection{Theoretical results on the benchmarks}

Here we review very briefly the known theoretical results on the related benchmarks. For both \DBV and \hottopic, the \oea is very well understood. For this simple algorithm, the main parameter is the mutation rate. For both benchmarks, there is a sharp phase transition in the mutation rate $\chi/N$. There is a constant $\chi_0$, which is $\chi_0 = 2.13\ldots$ for \DBV~\cite{lengler2021runtime} and $\chi_0 = 1.59..$ for \hottopic~\cite{lengler2018drift,lengler2019general}. If $\chi < \chi_0$ then the \oea finds the optimum in time $O(N\log N)$, otherwise it needs exponential time in $N$. The same dichotomy also holds for other dynamic linear functions, where all weights are drawn independently and identically distributed. The threshold $\chi_0$ depends on the distribution and is always larger than the threshold $1.59..$ for \DBV. For example, the threshold is $\chi_0 = 2$ for an exponential distribution and $\chi_0 = (2-p)/(1-p)$ for a geometric distribution with parameter $p$.

For \hottopic, it is known that this result generalizes to a large number of functions, including the \olea, the \moea, the \folea, the \fmoea, all without crossover, and the \ollga, where for the last three the parameter $\chi$ must be substituted appropriately~\cite{lengler2019general}. However, all these results come with an important caveat: they hold only sufficiently close to the optimum. This may seem like a harmless condition as usually the hardest part of optimization is close to the optimum. However, this is not generally the case for \hottopic. It was shown in~\cite{lengler2019exponential} that for \moea without crossover, the hardest part of \hottopic is not around the optimum, but that the algorithm instead gets stuck in a region ``in distance $\eps n$'' from the optimum, even though it would efficiently find the optimum if started in distance $\eps n/2$ from the optimum~\cite{lengler2019exponential}. This shows abstractly the surprising fact that progress from $\eps n$ to $\eps n/2$ is harder than from $\eps n/2$ to $0$. However, the $\eps$ could not be explicitly specified. One advantage of our benchmarking approach is that we are able to replace the abstract value of $\eps$ by some concrete numbers, see Section~\ref{sec:results}. 

Another interesting result on \hottopic is that crossover helps dramatically close to the optimum: for every mutation rate $\chi$ there is $\mu_0$ such that the \moga using either mutation or crossover, each with probability $0.5$, is efficient close to the optimum. It remained an open question how the behavior is further away from the optimum, where there are the opposing effects of being (counter-intuitively) in a harder region of the search space, and having the benefits of crossover.

For \DBV, results other than for the \oea are a bit more sparse. The benchmark has been introduced in~\cite{lengler2022large}. Experimental results for a very limited set of parameters indicated that also here the hardest part of optimization is not always next to the optimum, at least not for the $(2+1)$ EA and $(3+1)$ EA without crossover. For the $(2+1)$ EA this could also be shown formally in~\cite{lengler2021runtime}. However, it can be ruled out that the reason is the same as for the \moea on \hottopic, since the latter effect crucially depends on the fact that \hottopic functions maintain the order of weights for some period of time~\cite{lengler2022large}. So even though there is a similar surprising behavior, the reasons must be different.

\subsection{IOHprofiler}
IOHprofiler~\cite{doerr2019benchmarking} is a modular toolbox for benchmarking iterative optimization heuristics. It provides access to a variety of problem suites through a common interface in its IOHexperimenter~\cite{iohexp} module. This allows for a flexible benchmarking pipeline, which includes data logging. This data can be processed and visualized in the IOHanalyzer~\cite{IOHanalyzer} module.

By providing access to a wide variety of problem types with common logging infrastructure, IOHprofiler has enabled a range of different research questions, both from a theoretical and practical perspective. For example, recent works have shown that the star-discrepancy optimization problem provides a challenging environment for a large set of optimization heuristics~\cite{clement2023computing}, or created environments for competitions on several sets of submodular optimization problems~\cite{submodular}. 

\section{Experimental Setup}\label{sec:setup}
\subsection{DBV in IOHexperimenter} 

Our experimental setup is built upon the integration of several versions of the Dynamic BinVal problem class into the IOHexperimenter environment.
To handle the dynamic nature of the problem, a way of progressing its internal state (importance of each variable), a \textit{step}-function has to be added to the problem itself. Note that this means that in practice, the algorithm determines when the problem's state is updated, which is needed since the problem itself has no ability to detect when an evaluation belongs to a new generation. 

The Dynamic BinVal problem is based on the Binary Value problem defined via BinVal$:\{0,1\}^n \to \mathbb{R}; x\mapsto \sum_{i=1}^{n}2^{i-1}x_i$. In the dynamic version, we draw a permutation $\pi$ for each generation and compute all fitnesses with respect to the permuted weights $\DBV(x) = \DBV_{\pi}(x) = \sum_{i=1}^n 2^{\pi(i)-1}x_i$. This version of the problem can not be implemented directly, as the weights would become too large for the standard datatypes used within IOHexperimenter. As such, we follow the observation from~\cite{lengler2021runtime} and draw weights independently from different distributions instead. In this context, we consider three distinct distributions:
\begin{itemize}
    \item \textbf{Uniform}: Uniform Weights from $\mathcal{U}(0,1)$
    \item \textbf{PowersOfTwo:} Uniformly from the set $\{2^{i} : 1\le i \le 31-\log_2(N)\}$ (to avoid overflow of the summed value)
    \item \textbf{Pareto}: Pareto distribution with a limited upper bound ($(1-\mathcal{U}(0,0.75))^{-10}$, where 0.75 is chosen to avoid overflow of the summed value)
\end{itemize}

In addition to these 3 problem variants, we can also consider a 'true' version of the DBV (\textbf{Ranking}) by not implementing an evaluation, but a ranking function instead. This ranking function sorts the given individuals lexicographically according to the order of the weights, and thus does not require using the large powers of 2 directly. While this version is true to the problem, it requires any used optimization algorithm to be modified from objective-value evaluations to this ranking scheme, and thus does not give good modularity. 

The integration of the 4 versions of DBV into IOHexperimenter has the additional benefit that we can utilize several transformation methods to validate algorithmic invariances. These transformations, originally proposed in the context of the 'PBO' suite, include changing the target string with a random bitstring, swapping the order of variables and translating/scaling the returned objective value. In our experiments, when we refer to a problem instance, this corresponds to a function combined with a setting of these transformations.

\subsection{Used Algorithm}

In this paper, we make use of a standard GA which is heavily parameterized to allow for a variety of experimental setups. An outline of this algorithm is given in Algorithm~\ref{alg:GA}, while the available parameters are indicated in Table~\ref{tab:parameters}. For our termination criterion, we use a combination of function evaluation budget ($100 \cdot N$ by default) and optimality (to save computation time). 

\begin{algorithm}[t]
    \caption{Outline of the used Genetic Algorithm.}
    \label{alg:GA}
    \begin{algorithmic}[1]
        \State{Inputs: Parameters from Table~\ref{tab:parameters}, function $f$ to maximize}
    \Procedure{\textsc{GA}}{}
    \State{Initialize Population $\mathcal{P}$} \Comment{At random or with fixed distance to optimum}
        \While{termination criterion not met}
            \If{Update required} \Comment{Based on update frequency}
                \State{$f\rightarrow\textit{step}()$}
            \EndIf
            \For{$i$ in number of offspring}
                \State{Perform crossover with probability $p_c$, creating $\mathcal{O}_i$} \Comment{Uniformly}
                \If{No crossover or Mutation after cross disabled}
                    \State{$n_{\textit{flip}} \leftarrow max(n_{\textit{min}}, \textit{Binom}(\frac{\chi}{N}, N))$}
                    \State{Flip $n_{\textit{flip}}$ bits to create $\mathcal{O}_i$}
                \EndIf
            \EndFor
        \State{Evaluate $\mathcal{O}$ (and $\mathcal{P}$ if required)} \Comment{Rank or evaluate based on function}
        \State{$\mathcal{P} \leftarrow (\mathcal{P}; \mathcal{O})$} \Comment{Plus or comma selection}
    \EndWhile
    \EndProcedure
    \end{algorithmic}
\end{algorithm} 

\begin{table*}[]
    \centering
    \begin{tabular}{l | c | c | c }
        Parameter & Range & Default & Notes\\
        \hline
        Mutation factor $\chi$ & $[0,N]$ & $1$ & Mutation rate is $\frac{\chi}{N}$ \\
        Number of offspring $\lambda$ & $[1, \infty)$  & 1 & \\
        Number of parents $\mu$ & $[1, \infty)$ & 1 & \\
        Selection & plus, comma & plus & \\
        Crossover rate $p_c$ & $[0,1]$ & 0 & \\
        Dynamic frequency & $[0, \infty)$ & 1 & Number of generations   \\
        Minimum bits flipped $n_{\textit{min}}$ & $[0,N)$ &  0 &  \\
        Mutation after crossover & yes, no & yes & Mutate crossover-result \\
        \hline
        Function Version & \begin{tabular}{@{}c@{}}Rank, Uniform \\ Power2, Pareto\end{tabular} & Rank & Rank=lexicographic  \\
        Instance & $[0,\infty]$ & $[1,25]$ & Determines transformations \\
        Dimensionality $N$ & $[1,\infty]$ & 1000 & \\
    \end{tabular}
    \caption{Parameters available in the used GA version (top) and parameterization of the \DBV functions (bottom). In our experiments, we use the default value unless stated otherwise. }
    \label{tab:parameters}
\end{table*}

\subsection{Experimental Setup and Reproducibility}

Throughout our experiments, we make use of $N=1000$, and perform independent runs on $15$ different instances for each configuration. Each run is given a budget of $1000\cdot N$ evaluations. Our full experimental setup, including the data collection, processing and visualization, is available on our Zenodo repository~\cite{reproducibility_and_figures}. This repository also includes the full datasets from our experiments in an IOHanalyzer-compatible format.

\section{Results}\label{sec:results}

\subsection{Exploration of Used Algorithms}

\begin{figure}
    \centering
    \includegraphics[width=0.9\textwidth]{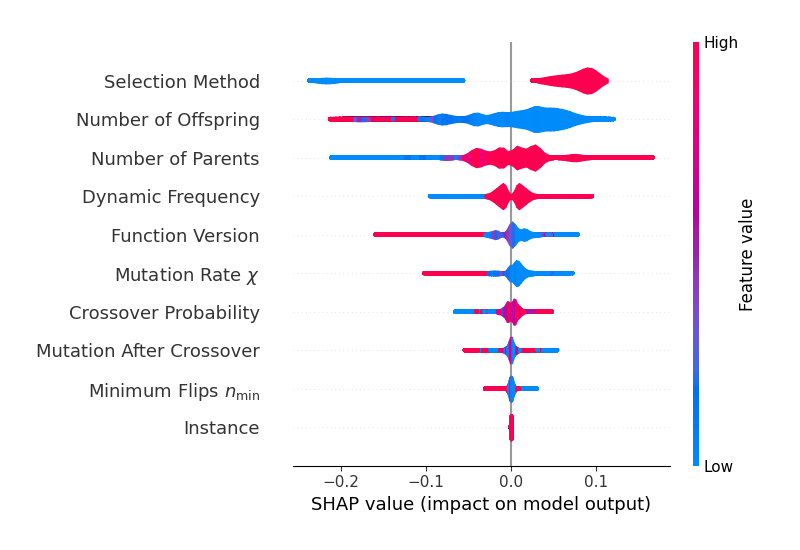}
    \caption{Dimenionality 1000, SHAP-values for each of the varied parameter settings. For selection, red corresponds to plus-selection. For the function version, red corresponds to the rank-based version. }
    \label{fig:explainability_overview}
\end{figure}

For our first set of experiments, we perform a coarse grid sampling of our parameter space to identify which factors most impact the algorithm's performance. For this setting, performance is measured as the percentage of correct bits after $100\cdot N$ evaluations. To analyze the resulting data, we make use of the SHAP approach~\cite{shap}, which is a popular approach in the context of global explainability~\cite{van2024explainable}. The resulting SHAP-values are shown in Figure~\ref{fig:explainability_overview}, sorted from most impactful (top) to least (bottom).
In this figure, the color of each dot corresponds to the chosen option, e.g. for selection blue corresponds to a comma while red corresponds to a plus-strategy.

While the grid used to create Figure~\ref{fig:explainability_overview} is rather coarse, it does provide an initial overview of the components of the algorithm which influence the selected performance measure in the most drastic ways.  In particular, we notice that using a comma selection greatly deteriorates the ability to get close the the optimum, and that using a low number of parents and a large number of offspring is beneficial in this experimental setting. This generalizes the theoretical result in~\cite{lengler2019exponential}, where it was identified as a problem of the \moea that the population is exchanged only slowly, and the family trees within the population become deep. Those properties are generally mitigated if the ratio of offspring versus parents is large. Also of note is the fact that the version of the function has quite some impact on the algorithm's performance, while the instantiation of the function is negligible. This suggests that the used GA is invariant to the used instance transformations, but not to the used weight-distribution.

\subsection{Mutation in 1+1}

For our second experiment, we focus on the (1+1) GA and study the impact of mutation and the used function version on its performance. To achieve this, we vary the mutation rate between $0$ and $6$ (with increments of $0.1$), as well as the minimum number of mutated bits between $0$ and $1$. This is repeated for each version of the \DBV, with $25$ independent runs of budget $1000\cdot N$.

\begin{figure}
    \centering
    \includegraphics[width=\textwidth]{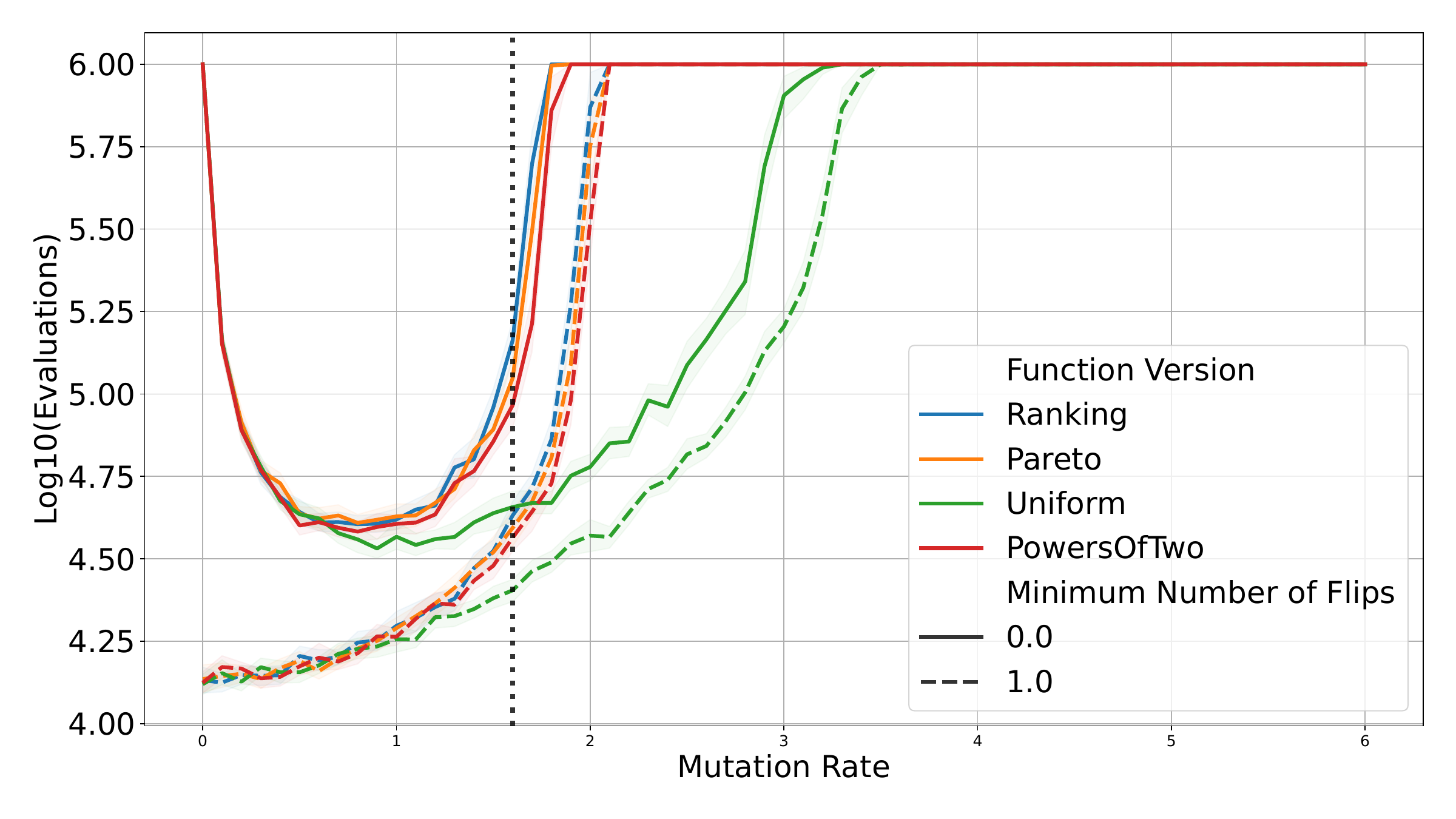}
    \caption{Dimensionality 1000, number of evaluations (log-scaled) needed to reach optimum given different mutation rates. }
    \label{fig:mutation_evals_1plus1}
\end{figure}

\begin{figure}
    \centering
    \includegraphics[width=\textwidth]{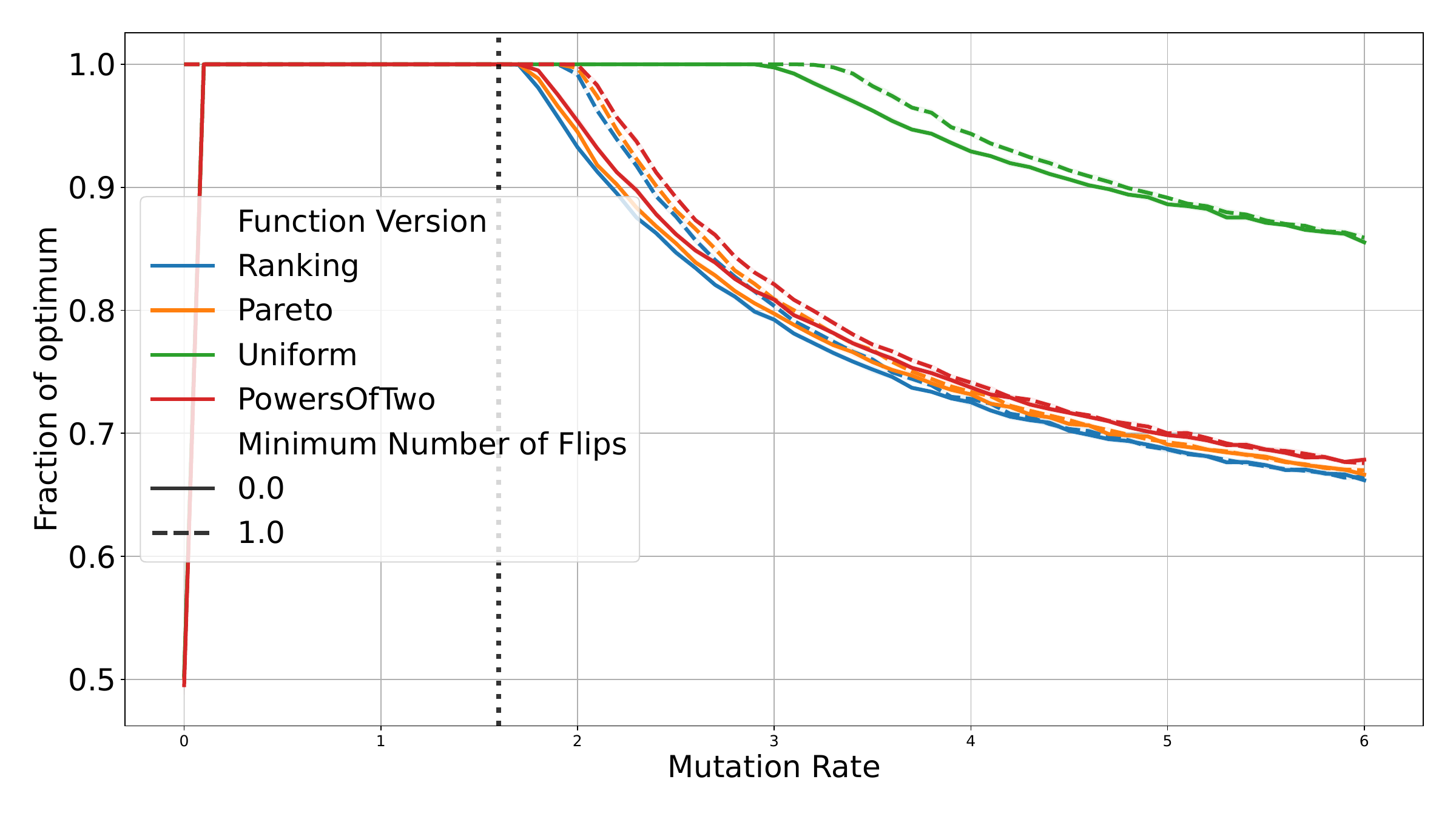}
    \caption{Dimensionality 1000, relative number of correct bits after the budget (1000 times $N$) is used, given different mutation rates. }
    \label{fig:mutation_fraction_1plus1}
\end{figure}

In Figure~\ref{fig:mutation_evals_1plus1}, we show the number of function evaluations required to reach the optimum for each of the selected GA versions. In Figure~\ref{fig:mutation_fraction_1plus1} we plot the fraction of correct bits after the evaluation budget has been exhausted. A dotted black line highlights the mutation rate of $1.6$, which is the theoretical threshold for the ranking-based version, the solid blue line, in the case where it is allowed that mutation flips zero bits (the parent is duplicated). Also, theory predicts that there are no other versions with smaller thresholds~\cite {lengler2018one}. Indeed, both theoretical predictions are confirmed in Figures~\ref{fig:mutation_evals_1plus1} and~\ref{fig:mutation_fraction_1plus1}. 

When comparing the different versions of \DBV in Figure~\ref{fig:mutation_evals_1plus1}, we observe quite a stark difference in algorithm performance between the Uniform-based \DBV and the other settings. While for low mutation rates all considered versions result in the same behavior, which results from flipping 1 bit, which is always accepted if it was a zero-bit and rejected otherwise, with larger mutation rates the uniform problem becomes visibly easier than the rest. This is visible both in the number of evaluations required to reach optimality, as well as in the number of correct bits after the budget threshold is reached (Figure~\ref{fig:mutation_fraction_1plus1}). We also see that both the PowersOfTwo and Pareto are very similar to the Rank-based \DBV (with a slight preference for Pareto), so they seem to both be suitable to model the problem when function-evaluation-based approaches are required.

To understand the dashed lines, we observe that in the $(1+1)$ case, generations have no effect if the parent is duplicated (except for counting as an idle step). Hence, we should consider the number $n_{\text{flip}}$ of flipped bits \emph{conditioned} on flipping at least one bit. A bit counter-intuitively, the expectation $\mathbb{E}[n_{\text{flip}} \mid n_{\text{flip}} >0]$ goes \emph{down} by overwriting the case $n_{\text{flip}} =0$ with $n_{\text{flip}}=1$, which means that this change \emph{decreases} the number of flipped bits in the conditional space. Effectively this change decreases the mutation rate, which is why the threshold is shifted to the right.

Figure~\ref{fig:mutation_fraction_1plus1} shows that the effect of a large mutation rate is quite dramatic. For example, for a mutation rate of $3/n$ the algorithm is only able to set less than $80\%$ of the bits correctly after a very generous budget of $1000N$. This is still a large distance from the optimum. Even with mutation rate of $3/n$, the algorithm has a chance of roughly $15\%$ (for $n_{\min}=0$) to flip exactly one bit in a mutation, compared to $37\%$ for mutation rate $1/n$. Since those are the mutations that bring most progress in this range, the poor performance can not be explained by the moderate slowdown (factor $2$-$3$) that larger mutation rates entail on other benchmarks like \onemax. The large effect shown in Figure~\ref{fig:mutation_fraction_1plus1} has not been quantified by theory before.

\begin{figure}
    \centering
    \includegraphics[width=0.9\textwidth]{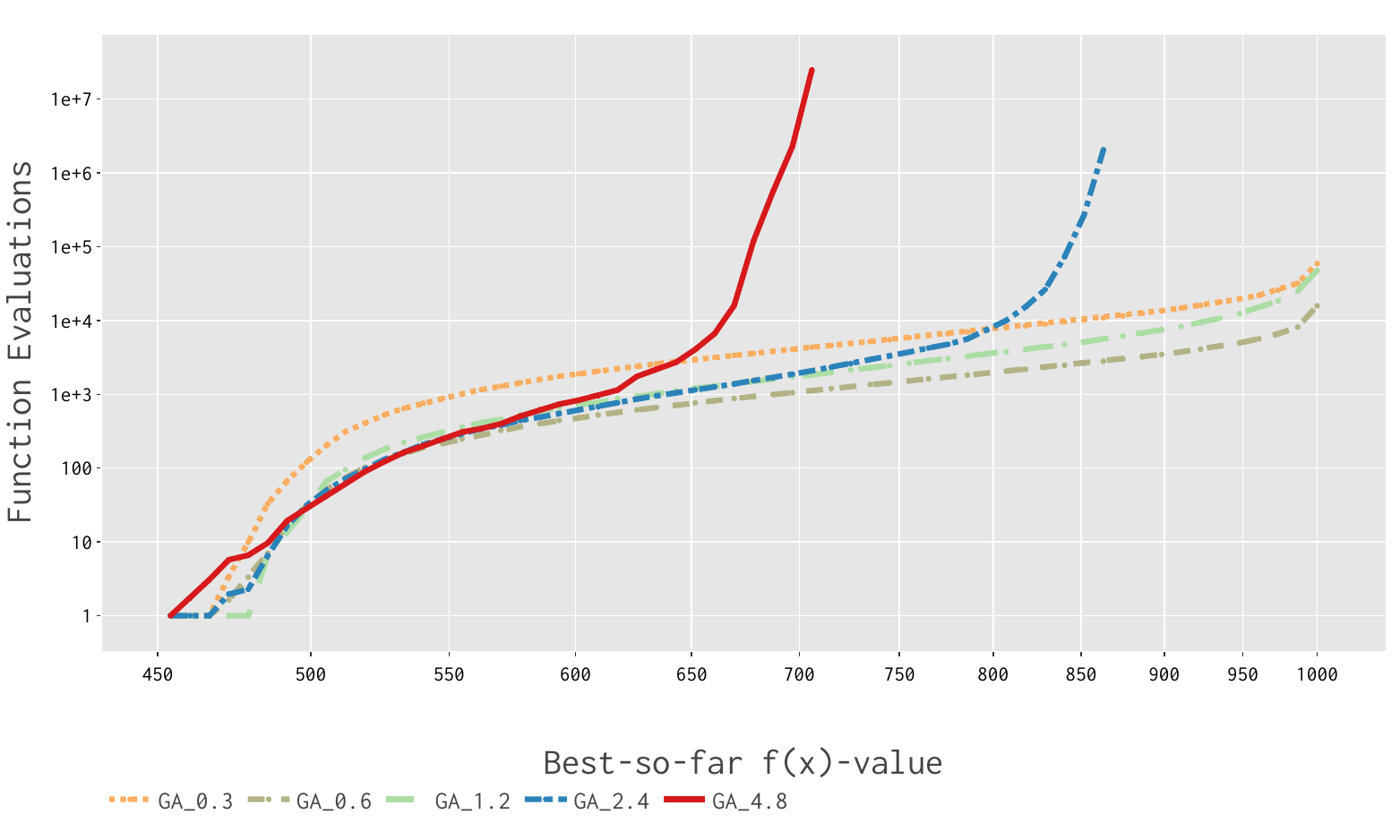}
    \caption{Dimensionality 1000, some selected ERT curves for different mutation rates (rank-based function). }
    \label{fig:ert}
\end{figure}

While the heavy aggregation of Figures~\ref{fig:mutation_evals_1plus1} and \ref{fig:mutation_fraction_1plus1} shows the final performance of the considered algorithm configurations, it does not show the full behavior of the underlying runs. Since we recorded the full performance trajectory, we can load this data into IOHanalyzer to generate more fine-grained visualization, such as ERT curves~\cite{IOHanalyzer} illustrated in Figure~\ref{fig:ert}. The ERT gives some estimates for the expected running time under some pessimistic assumptions on the running time in which the optimum is not reached~\cite{IOHanalyzer}. It tends to overestimate the actual running times~\cite{lengler2022large}. Again, this shows how suddenly the performance drops at a given level, and how little is to be gained by increasing the computation budget.

\subsection{Adding multiple parents: Crossover}

For our next set of experiments, we investigate the effects of crossover (uniform) on the performance of the $(\mu+1)$-GA. We vary $\mu$ in $\{2,3,4,8,16\}$, and crossover rate in $\{0,0.5,0.9\}$. In addition to the crossover rate, we change whether an individual resulting from crossover still undergoes mutation or not. Mutation is not applied after crossover in the previous theory results.

\begin{figure}
    \centering
    \includegraphics[width=\textwidth]{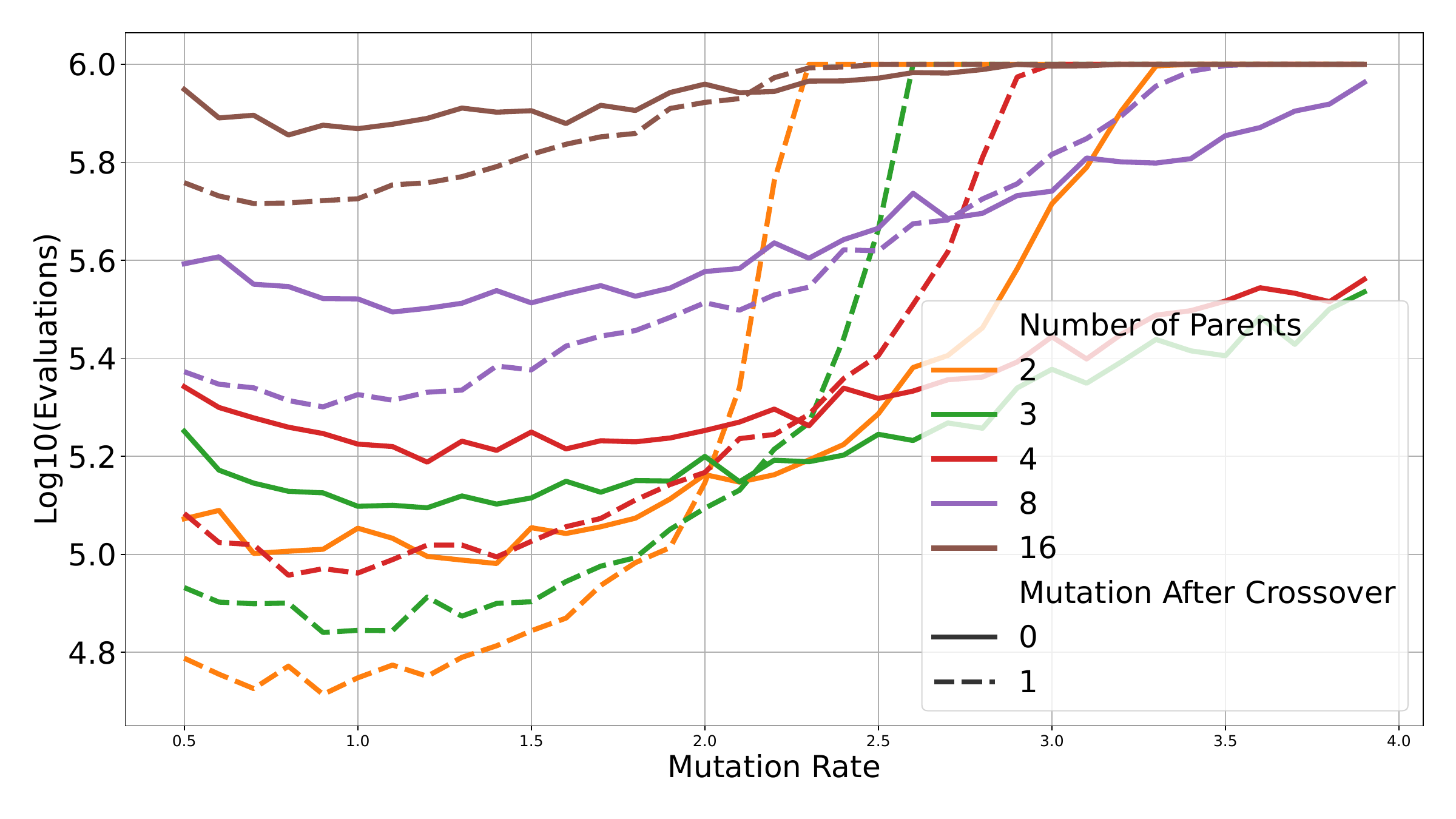}
    \caption{Dimensionality 1000, number of evaluations (log-scaled) needed to reach optimum, on the rank-based functions, given different number of parents and mutation after crossover setting. Crossover probability is $0.5$. }
    \label{fig:mut_after_cross}
\end{figure}

To illustrate the impact of the different versions of crossover, we plot the number of evaluations until optimality for a given crossover rate $0.5$ in Figure~\ref{fig:mut_after_cross}. In this figure, we can see a transition between these two mechanisms, where for lower mutation rates it is beneficial to still perform mutation after crossover, while disabling mutation once crossover is performed allows for more stable behavior at higher mutation rates. One perspective on this is that there is something like an optimal amount of mutation. If the mutation rate is smaller, then it may be beneficial to add mutation also after crossover. If the mutation rate is larger, then it is beneficial to not apply it in every generation.

A similar trade-off can be observed in Figure~\ref{fig:mut_after_cross} for population sizes. While an increase in population size requires more function evaluations per iteration (each individual needs to be re-evaluated to perform selection), it also improves the ability of the algorithm to work with much higher mutation rates. 

\subsection{Difficulty of the Search Stages}

\begin{figure}
    \centering
    \includegraphics[width=\textwidth]{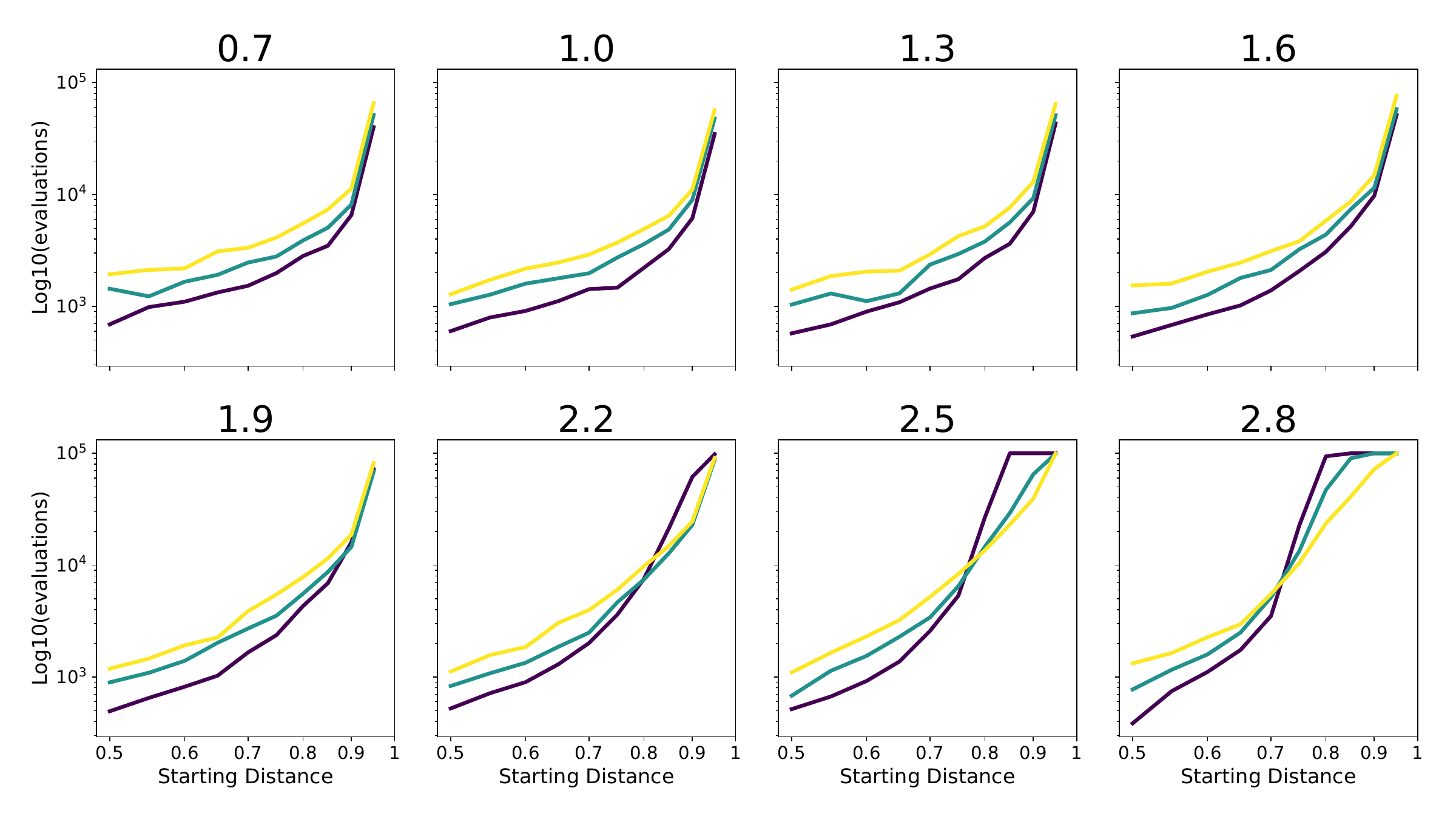}
    \caption{Dimensionality 1000, number of evaluations (log-scaled) needed to improve the hamming distance to the optimum solution by 50, on the rank-based functions, given different number of parents (blue = 2, green = 3, yellow = 4) and mutation rate (plot titles). Crossover probability is $0$.}
    \label{fig:starting_dist_plot2}
\end{figure}

A remarkable theoretical result is that the hardest region for optimization is not always around the optimum. This has been shown theoretically for the $(2+1)$ EA on \DBV~\cite{lengler2021runtime} and for the $(\mu+1)$ EA with large constant $\mu$ on \hottopic~\cite{lengler2019exponential}. Notably, the underlying cause is very different in both settings. In this section we want to investigate how relevant this phenomenon for \DBV in practice, which has also been shown in some narrow-focused experiments in~\cite{lengler2022large}, for example for the $(3+1)$ EA for mutation rate $2.7/N$. To this end, we seeded the algorithm with bit strings of a given Hamming distance from the optimum (fraction of correct bits). By seeding the algorithm with strings ranging from $0.5$ to $0.95$ and measuring the time needed to improve this fraction by $0.05$, we get an overview of the relative hardness of each stage of the optimization procedure. The results of this experiment are visualized in Figure~\ref{fig:starting_dist_plot2}. 

In our experiments, we could not reproduce the finding that the hardest region was in some intermediate range. Comparing with the experiments in~\cite{lengler2022large}, we suspect that this effect only shows for very large budgets. Since we want to systematically explore different parameter setting in this paper, the required budget becomes forbiddingly large. We believe our results do show that the effect may not be very common in practical setting, and we leave it to future work to nail down the difference further. However, in Figure~\ref{fig:starting_dist_plot2} we see another interesting effect. As for the mutation rate, the number of parents has a trade-off between different regions of the search space. We generally see that larger population sizes are beneficial in ``easy'' regions of the search space, but are detrimental in ``harder regions''. This suggests that in the setting of dynamic optimization and \DBV, there may not be a fixed population size that is optimal throughout all phases of the optimization process.

\subsection{HotTopic: impact of update frequency}

While the DBV problem is defined as a fully dynamic problem where the environment changes every iteration of the algorithm, we may also change it less frequently. This has been conjectured to be related to the well-studied \hottopic problem, which is a \emph{static} monotonic problem, but where different regions of the search space behave like different linear functions.

\begin{figure}[t]
    \centering
    \includegraphics[width=\textwidth]{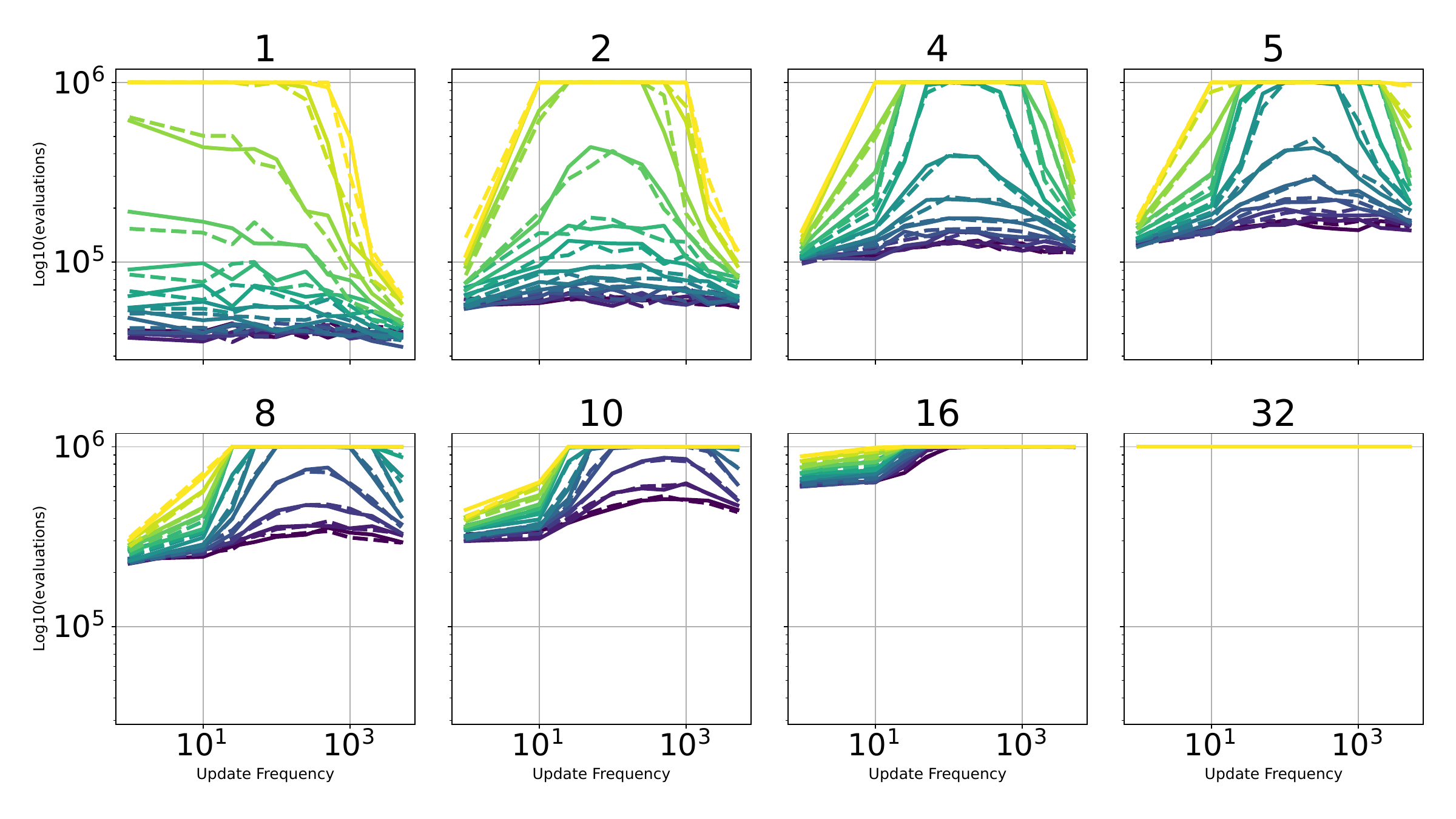}
    \caption{Dimensionality 1000, number of evaluations (log-scaled) needed to reach optimum, on the rank-based functions, given different number of parents, mutation rates and mutation after crossover setting ($p_c=0.9$). The x-axis corresponds to the frequency at which the weights are updated (in number of generations)}
    \label{fig:hottopic}
\end{figure}

To illustrate the impact of the function update frequency, we again show the number of evaluations required by our $(\mu, 1)$-GA with different settings of $\mu$. The results of this experiment, where the crossover rate is $0.9$ and mutation after crossover is enabled, can be seen in Figure~\ref{fig:hottopic}. From this figure, we observe that for $\mu>1$ the difficulty of the problem increases with the update frequency up until some point, after which the problem difficulty drops again. With a large enough update frequency, the problem becomes essentially static, at which point the problem collapses to a static binary value problem. While this function is by folklore conjectured to be the hardest static linear function, the class of static linear functions is rather easy, and this includes the static binary value problem. 

Interestingly, the intermediate range where the problem is hardest may correspond to \hottopic functions, where informally the ``environment'' changes after $\eps N$ generations, where $\eps$ is a small constant. There is a remarkable analogy to the theoretical results in~\cite{lengler2019exponential}. There it was also shown that an intermediate regime is the hardest, but not for the update frequency, but rather for the distance from the optimum. However, the two quantities may be related in this context. In the \hottopic analysis, the crucial observation was that  if the weight structure is stable, then some bad family tree structure may evolve temporarily. However, in some distance from the optimum, improvements are found while this bad family tree structures dominate. Similarly, here the same bad family tree structures might evolve while the environment is constant, and new offspring may evolve while these bad structures are around. We hope that this hypothesis gives inspiration for future work, be it empirically or theoretically.

\section{Conclusions and Future Work}

Through a varied set of empirical experiments, we have highlighted several ways in which benchmarking can be used to build upon results from theoretical analysis in sometimes surprising ways. By integrating different versions of the Dynamic Binary Value problem into the IOHprofiler platform, we were able to compare several versions of \DBV, highlighting that sampling-based versions, specifically based on the Pareto distribution, lead to similar performance as the original \DBV formulation, without requiring impractically large weights. 

To highlight the potential of the integration of \DBV into different benchmarking pipelines, we analyzed the impact of a variety of algorithmic modifications to a standard GA, which allows for the exploration of research questions related to mutation rates, crossover methods and population sizes within the same setup. Perhaps most intriguing from a theory perspective is the non-monotone dependence on the update frequency in Figure~\ref{fig:hottopic}. For symmetry reasons, the update frequency makes no difference for population size $\mu=1$. Hence, the observation that a lower update frequency makes the problem harder must be due to the structure of the population. It would be very interesting to understand how the structure of the evolving populations depends on the update frequency, and why a lower update frequency leads to more susceptible populations.

The experimental environment described in this paper can be further extended to look at different kinds of dynamic environments or different algorithmic ideas. Such new empirical studies might be used to determine interesting aspects of these problems to further analyze theoretically, leading to a positive feedback loop between theory and practice. 

\vspace{1.5ex}
\textbf{Acknowledgements}
This work was initiated as part of the Dagstuhl Seminar 23332 "Synergizing Theory and Practice of Automated Algorithm Design for Optimization". Our work was supported by CNRS Sciences informatiques via the AAP project IOHprofiler.

%
%
%
\bibliographystyle{splncs04}
\bibliography{references}

\begin{thebibliography}{10}
\providecommand{\url}[1]{\texttt{#1}}
\providecommand{\urlprefix}{URL }
\providecommand{\doi}[1]{https://doi.org/#1}

\bibitem{branke2012evolutionary}
Branke, J.: Evolutionary optimization in dynamic environments, vol.~3. Springer Science \& Business Media (2012)

\bibitem{clement2023computing}
Cl{\'e}ment, F., Vermetten, D., De~Nobel, J., Jesus, A.D., Paquete, L., Doerr, C.: Computing star discrepancies with numerical black-box optimization algorithms. In: Proceedings of the Genetic and Evolutionary Computation Conference. pp. 1330--1338 (2023)

\bibitem{doerr2019benchmarking}
Doerr, C., Ye, F., Horesh, N., Wang, H., Shir, O.M., B{\"a}ck, T.: Benchmarking discrete optimization heuristics with iohprofiler. In: Proceedings of the Genetic and Evolutionary Computation Conference Companion. pp. 1798--1806 (2019)

\bibitem{Janett2023two}
Janett, D., Lengler, J.: Two-dimensional drift analysis: Optimizing two functions simultaneously can be hard. Theoretical Computer Science  \textbf{971},  114072 (2023)

\bibitem{lengler2019general}
Lengler, J.: A general dichotomy of evolutionary algorithms on monotone functions. IEEE Transactions on Evolutionary Computation  \textbf{24}(6),  995--1009 (2019)

\bibitem{lengler2019does}
Lengler, J., Martinsson, A., Steger, A.: When does hillclimbing fail on monotone functions: an entropy compression argument. In: 2019 Proceedings of the Sixteenth Workshop on Analytic Algorithmics and Combinatorics (ANALCO). pp. 94--102. SIAM (2019)

\bibitem{lengler2022large}
Lengler, J., Meier, J.: Large population sizes and crossover help in dynamic environments. Natural Computing pp. 1--15 (2022)

\bibitem{lengler2021runtime}
Lengler, J., Riedi, S.: Runtime analysis of the {($\mu$+ 1)-EA} on the dynamic binval function. Evolutionary Computation in Combinatorial Optimization  \textbf{12692},  84--99 (2021)

\bibitem{lengler2018one}
Lengler, J., Schaller, U.: The {(1+1)-EA} on noisy linear functions with random positive weights. In: 2018 IEEE Symposium Series on Computational Intelligence (SSCI). pp. 712--719. IEEE (2018)

\bibitem{lengler2018drift}
Lengler, J., Steger, A.: Drift analysis and evolutionary algorithms revisited. Combinatorics, Probability and Computing  \textbf{27}(4),  643--666 (2018)

\bibitem{lengler2019exponential}
Lengler, J., Zou, X.: Exponential slowdown for larger populations: the ($\mu$+ 1)-ea on monotone functions. In: Proceedings of the 15th ACM/SIGEVO Conference on Foundations of Genetic Algorithms. pp. 87--101 (2019)

\bibitem{shap}
Lundberg, S.M., Lee, S.I.: A unified approach to interpreting model predictions. Advances in neural information processing systems  \textbf{30} (2017)

\bibitem{submodular}
Neumann, F., Neumann, A., Qian, C., Do, A.V., de~Nobel, J., Vermetten, D., Ahouei, S.S., Ye, F., Wang, H., B{\"{a}}ck, T.: Benchmarking algorithms for submodular optimization problems using iohprofiler. CoRR  \textbf{abs/2302.01464} (2023). \doi{10.48550/arXiv.2302.01464}, \url{https://doi.org/10.48550/arXiv.2302.01464}

\bibitem{neumann2020analysis}
Neumann, F., Pourhassan, M., Roostapour, V.: Analysis of evolutionary algorithms in dynamic and stochastic environments. Theory of evolutionary computation: recent developments in discrete optimization pp. 323--357 (2020)

\bibitem{nguyen2012evolutionary}
Nguyen, T.T., Yang, S., Branke, J.: Evolutionary dynamic optimization: A survey of the state of the art. Swarm and Evolutionary Computation  \textbf{6},  1--24 (2012)

\bibitem{iohexp}
de~Nobel, J., Ye, F., Vermetten, D., Wang, H., Doerr, C., B{\"{a}}ck, T.: {IOHexperimenter}: Benchmarking platform for iterative optimization heuristics. CoRR  \textbf{abs/2111.04077} (2021), \url{https://arxiv.org/abs/2111.04077}

\bibitem{van2024explainable}
van Stein, N., Vermetten, D., Kononova, A.V., B{\"a}ck, T.: Explainable benchmarking for iterative optimization heuristics. arXiv preprint arXiv:2401.17842  (2024)

\bibitem{reproducibility_and_figures}
Vermetten, D., Lengler, J., Rusin, D., B{\"a}ck, T., Doerr, C.: Reproducibility files and additional figures  (2024), code and data repository (Zenodo): \url{doi.org/10.5281/zenodo.10964455} Figure repository (Figshare): \url{doi.org/10.6084/m9.figshare.25592904}

\bibitem{IOHanalyzer}
Wang, H., Vermetten, D., Ye, F., Doerr, C., B{\"{a}}ck, T.: Iohanalyzer: Detailed performance analysis for iterative optimization heuristic. ACM Trans. Evol. Learn. Optim.  \textbf{2}(1),  3:1--3:29 (2022). \doi{10.1145/3510426}, \url{https://doi.org/10.1145/3510426}, iOHanalyzer is available at CRAN, on GitHub, and as web-based GUI, see \url{https://iohprofiler.github.io/IOHanalyzer/} for links

\end{thebibliography}

\end{document}